\documentclass[letterpaper, 10 pt, conference]{ieeeconf}  

\IEEEoverridecommandlockouts                              
\usepackage{graphics} 
\usepackage{epsfig} 
\usepackage{subfigure}
\usepackage{xcolor}
\usepackage{cite}

\title{\LARGE \bf
Tracker-Free Robotic Ultrasound Calibration with a Spherical-Marker
Phantom and Threshold-Free Center Localization}

\author{
Kyoungmo Koo$^{1,2,*}$,
Guangshen Ma$^{1,*, \ddagger}$,
Xueding Wang$^{2, \dagger}$,
and Mark Draelos$^{1,2,3, \dagger}$%
\thanks{*These authors contributed equally to this work. }%
\thanks{$^{\ddagger}$ Project Lead.}
\thanks{$^{\dagger}$Corresponding authors are Mark Draelos and Xueding Wang. Their email addresses are mdraelos@umich.edu and xdwang@umich.edu.}%
\thanks{$^{1}$Kyoungmo Koo, Guangshen Ma, and Mark Draelos are with the Department of Robotics, University of Michigan, 2505 Hayward St, Ann Arbor, MI 48109, USA. Their email addresses are kmkoo@umich.edu, guangshe@umich.edu, and mdraelos@umich.edu.}%
\thanks{$^{2}$Kyoungmo Koo, Xueding Wang, and Mark Draelos are with the Department of Biomedical Engineering, University of Michigan, 1150 Medical Center Dr, Ann Arbor, MI
48109, USA. Their email addresses are kmkoo@umich.edu, xdwang@umich.edu, and mdraelos@umich.edu.}%
\thanks{$^{3}$Mark Draelos is with the Department of Ophthalmology and Visual Sciences, University of Michigan Medical School, 1000 Wall St,
Ann Arbor, MI 48105, USA. His email address is mdraelos@umich.edu.}%
}

\usepackage{amsmath}
\usepackage{amssymb}
\usepackage{graphicx}

\begin{document}

\maketitle
\thispagestyle{empty}
\pagestyle{empty}

\begin{abstract}

Robotic ultrasound (US) calibration is essential
for accurately relating US images to the robot coordinate system, but
accurate and automated calibration remains challenging because existing
methods often require complex phantoms or external 3D trackers.
In this work, we develop a tracker-free robotic US calibration framework using a spherical-marker phantom and a highly automated perception pipeline for sphere-center localization.
The proposed threshold-free image-processing method localizes the spherical feature based on each US image's intensity distribution, eliminating hand-tuned intensity thresholds and improving robustness across imaging settings without per-system retuning.
Multi-pose observations of the spherical fiducial are then used to estimate the US-to-EE transformation without external tracking or prior localization of the sphere center in the robot base frame.
We validate the proposed framework on a robotic US platform through repeated sphere scans and further assess the calibrated system using geometrically distinct phantoms with known CAD models.
Across 12 cross-validation folds, single-marker calibration achieved sphere-center accuracy and precision of $1.75\pm0.51$~mm and $0.85\pm0.11$~mm, respectively, compared with $1.72\pm0.51$~mm and $0.84\pm0.12$~mm for three-marker calibration. Across the three reconstruction sets, the single-marker calibration yielded pooled post-registration point-to-surface MAE$\pm$SD values of $0.41\pm0.35$~mm for the cone and $0.50\pm0.41$~mm for the triangular prism, closely matching the three-marker results of $0.40\pm0.35$~mm and $0.48\pm0.40$~mm, respectively.

\end{abstract}

\section{Introduction}

\begin{figure}[t]
\centering
\includegraphics[width=0.90\columnwidth]{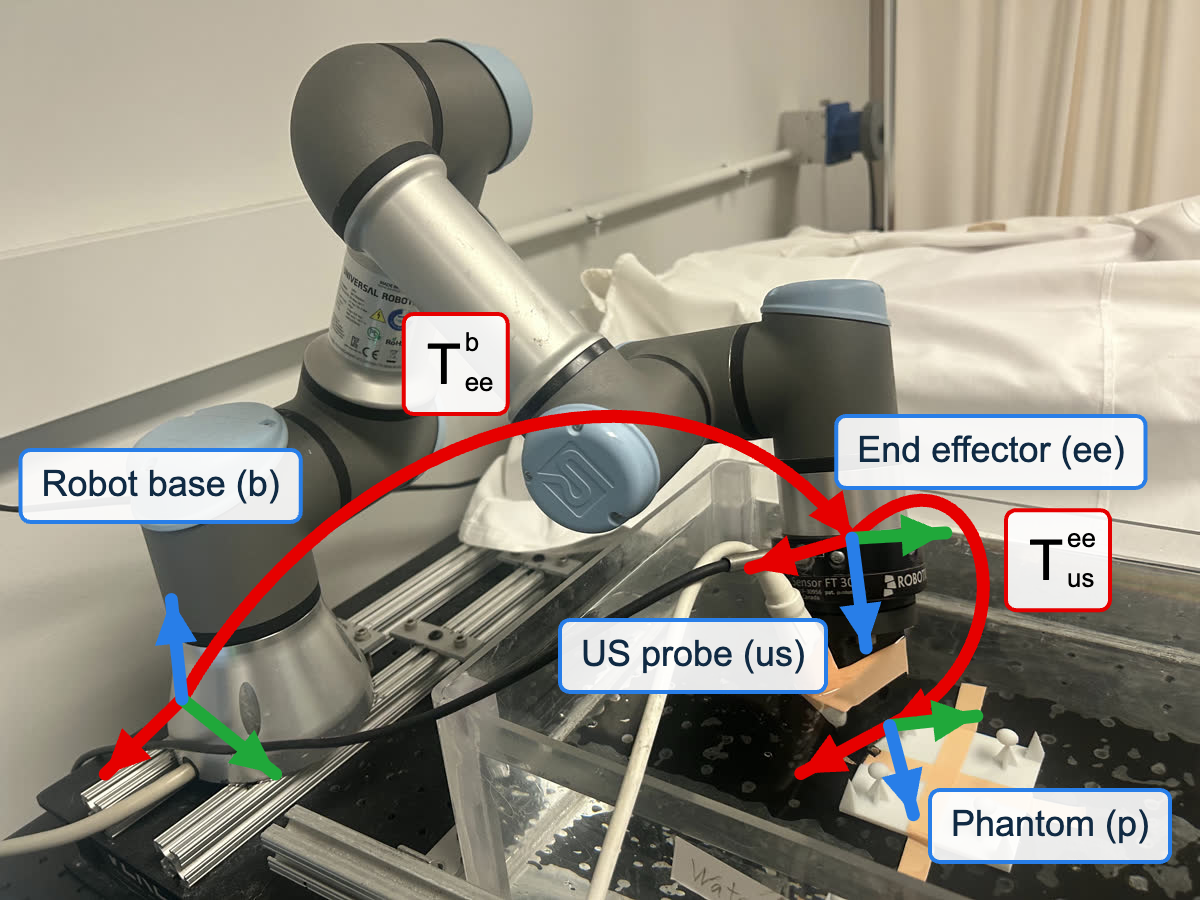}
\caption{Robotic ultrasound system scanning the sphere phantom in a water tank. The US probe is rigidly attached to the robot end-effector through a custom 3D-printed platform. The robot-base, end-effector, and US-image frames are related by the forward-kinematics transform $\mathbf{T}^{b}_{ee}$ (known) and the US-to-EE calibration $\mathbf{T}^{ee}_{us}$ (to be estimated).}
\label{fig:setup}
\end{figure}

In medical robotics, ultrasound (US) is widely used across clinical applications based on its real-time imaging capability, portability, low cost, and absence of ionizing radiation~\cite{sohn2023evidence, jiang2023robotic}, but its image quality and reproducibility remain highly operator-dependent~\cite{huang2023review,peng2025photoacoustic}.
The integration of US imaging with robotic systems can improve acquisition consistency through repeatable probe positioning, controlled scanning trajectories, and automated image acquisition~\cite{jiang2023robotic}, enabling applications such as 3D volumetric reconstruction~\cite{yang2021automatic,koo2025volumetric} and adaptive probe positioning for maintaining appropriate probe orientation and image quality over complex tissue surfaces~\cite{jiang2020automatic,koo2025volumetric}.
However, accurate spatial interpretation of the acquired US images requires determining the rigid transformation between the US image plane and the robot end-effector (EE).
This probe calibration is therefore a fundamental prerequisite for spatially accurate robotic US imaging~\cite{mercier2005review}, reconstruction, and control~(e.g., Fig.~\ref{fig:setup}).

Despite efforts to address this fundamental calibration requirement for robotic ultrasound systems, three key limitations persist across existing methods.
First, many existing methods rely on an external optical or electromagnetic (EM) tracker to recover probe poses during calibration~\cite{bo2015versatile, li2021framework, ryu2025rapid, liu2022handeye, shapeprior2025}. 
Such trackers are costly, lengthen the pipeline with two additional transformations (robot-to-tracker and tracker-to-probe), and impose line-of-sight or EM-distortion constraints on deployment. 
Second, a few methods eliminate external tracking by recovering probe poses directly from the robot's forward kinematics; however, these approaches have primarily been demonstrated with line-, cone-, or active-point targets~\cite{oliveira2025calibration, yang2025robot, aalamifar2016robot}. These targets have distinct limitations. A wire target allows the reconstructed point to slide freely along the wire, resulting in a structural loss of calibration information in that direction. A cone target is reliably visible only near its apex and within a limited range of viewing angles, making it difficult to acquire observations from diverse probe orientations. An active-echo target requires dedicated hardware components and synchronization with the US system, which increases system complexity and limits applicability to general imaging platforms.
These limitations can increase the complexity of target design, data acquisition, or calibration formulation, motivating a simpler approach based on repeated observations of a single passive fiducial.
Third, target detection is commonly automated through a fixed intensity threshold that must be retuned across machines, gain settings, and imaging depth~\cite{li2021framework, ryu2025rapid}, limiting reproducibility and cross-system generalization. 

To address these limitations, we present a tracker-free robotic US calibration framework using multi-pose observations of a single spherical fiducial and robot forward kinematics, without using  an external 3D tracker or a complex calibration phantom. Furthermore, a sphere requires neither the frame assembly and wire tensioning of wire phantoms nor the sharp, precisely fabricated apex of a cone.
The proposed framework combines automated, threshold-free sphere localization with robot-based pose estimation to simplify the calibration procedure.
The contributions are summarized as: 
\begin{enumerate}
\item \textbf{Tracker-free robot-US calibration using spherical fiducials:} We formulate a calibration framework that incorporates robot forward kinematics and sphere-based geometric constraints, eliminating external 3D tracking and prior localization of the fiducial centers in the robot base frame. 
The single-marker calibration achieved a sphere-center accuracy of
$1.75\pm0.51$~mm.
\item \textbf{Threshold-free sphere-center localization:} We develop an efficient localization method that estimates the sphere center from the intensity distribution of the acquired US images, eliminating manually selected intensity thresholds and per-system threshold tuning.
\item \textbf{Single- versus three-marker generalization comparison:} We evaluate calibration precision and accuracy on a physical robotic US platform through repeated spherical-fiducial scans, including a direct comparison between single- and three-sphere calibration under the same robot kinematics. 
\end{enumerate}

\section{Related Work}
\label{sec:related}

\subsection{Conventional Robot--US Calibration}

Robotic US calibration estimates the rigid transformation between the US image and robot end-effector frames~\cite{aalamifar2016robot,jiang2023robotic}.
Conventional calibration approaches typically establish this transformation through observations of known geometric features while obtaining the corresponding probe poses from either an external tracking system or the robot's forward kinematics. 
The calibration accuracy therefore depends not only on the geometric formulation, but also on how reliably the probe pose and calibration features can be localized. 
Existing methods can be broadly categorized into tracker-based approaches~\cite{
bo2015versatile,li2021framework,ryu2025rapid,
liu2022handeye,paralikar2025} and tracker-free approaches that directly exploit robot kinematics~\cite{
oliveira2025calibration,yang2025robot, aalamifar2016robot}.

\subsection{Robot--US Calibration with an External Tracker}
Most calibration methods recover the probe pose from an external optical or electromagnetic (EM) tracker. 
B\o{} et al. move a plastic sphere through the image plane with a UR5 robot while an optical tracker supplies the probe pose~\cite{bo2015versatile}. 
Li et al. use an optically tracked 3D-printed sphere with one-time pointer registration~\cite{li2021framework}, while Ryu et al. employ an optically tracked CT-visible multi-sphere phantom~\cite{ryu2025rapid}.
Wire- and N-wire--based phantoms with optical or EM tracking are also widely used~\cite{liu2022handeye,shapeprior2025}, while Paralikar et al. use an EM-tracked conical-tip stylus with a KUKA robot~\cite{paralikar2025}.

Although external tracking provides an accurate spatial reference for calibration, it introduces additional hardware requirements and system-level complexity~\cite{liu2022handeye,yang2025robot}.
The external tracker must provide sufficient precision for the desired calibration performance, often requiring specialized and costly equipment~\cite{li2021framework,liu2022handeye}.
External tracking introduces additional robot-to-tracker and tracker-to-probe transformations, increasing calibration complexity~\cite{bo2015versatile,li2021framework,liu2022handeye}.
Optical line-of-sight constraints and EM field distortion further limit deployment~\cite{sorriento2019optical}, motivating tracker-free approaches that instead exploit robot kinematics~\cite{aalamifar2016robot,oliveira2025calibration,yang2025robot}.

\subsection{Tracker-Free Calibration}

Tracker-free methods instead exploit robot forward kinematics, eliminating external tracking and its associated transformations.
Oliveira et al.~\cite{oliveira2025calibration} estimate the US-to-EE transformation by scanning an unknown straight-wire phantom and enforcing straightness of the reconstructed wires, with accuracy independently evaluated using a robot-mounted, pivot-calibrated needle and a 3D-printed cone.
Yang et al.~\cite{yang2025robot} register detected tips of four 3D-printed cones to a known model, while Aalamifar et al.~\cite{aalamifar2016robot} use an active-point fiducial for tracker-free calibration.

While these studies demonstrate the feasibility of tracker-free calibration using robot kinematics, existing approaches remain dependent on specific target geometries or instrumentation.
Recent methods use multi-cone point registration~\cite{yang2025robot} or single-cone conic-section geometry~\cite{jiang2026simplifying}, while active-point approaches require additional instrumentation~\cite{cheng2017active}.
Most closely related, Li et al.~\cite{li2021framework} use a spherical phantom for automatic robotic US calibration, but require camera-to-robot calibration and parameter-dependent sphere segmentation.
These limitations motivate our use of a passive spherical fiducial, whose rotational symmetry supports pose-diverse observations, together with robot kinematics and threshold-free center localization.

\subsection{Automated Target Localization of Ultrasound Images}
Across both tracker-based and tracker-free calibration pipelines, target localization commonly relies on fixed intensity thresholds or manually tuned detection parameters~\cite{li2021framework, ryu2025rapid}.
Such parameters may require adjustment across US systems, gain settings, and imaging depths, limiting reproducibility and cross-system generalization.
These limitations motivate an automated localization strategy that adapts to the intensity characteristics of each acquired image without manually specified intensity thresholds, motivating the method to localize the spherical fiducial from the image's own intensity distribution, requiring no per-system threshold tuning.

\section{Methods}
\label{sec:method}

\begin{figure*}[t]
\centering
\includegraphics[width=0.90\textwidth]{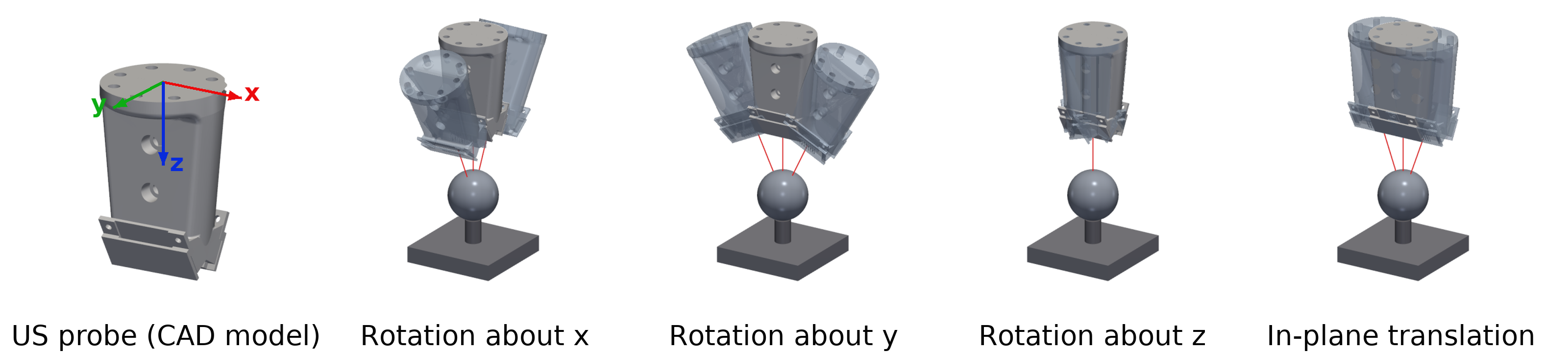}
\caption{
Scan trajectory about a single sphere. The leftmost panel shows the US probe-holder CAD model and coordinate frame, with $x$ and $y$ in the EE plane and $z$ along the axial direction. 
The probe is rotated by $\pm10^\circ$ about each axis and translated in-plane by $\pm3$~mm, yielding $45$ poses from nine orientations and five translations. 
Red lines indicate the line of sight to the sphere center.
}
\label{fig:traj}
\end{figure*}

\subsection{System Overview and Calibration Formulation}
We evaluate the proposed calibration framework on a UR3 robotic arm (Universal Robots, Odense, Denmark) carrying an L8-18i-D high-frequency linear-array probe driven by a GE Vivid E95 ultrasound system (GE Healthcare, Chicago, IL, USA).
A custom 3D-printed holder rigidly mounts the US probe to the robot end-effector.

Specifically, four custom 3D-printed components are used in the study. 
The calibration phantom contains four spherical ultrasound targets of radius $5$~mm at the corners of a $50$-mm square. 
Four surrounding cone vertices serve as touch targets and lie in the same horizontal plane as the sphere centers (Fig.~\ref{fig:process}(a)). 
The known sphere-center geometry is used in the three-marker calibration objective.
The proposed framework estimates the rigid transformation between the US image and robot end-effector frames using multi-pose observations of a spherical fiducial and robot forward kinematics, without requiring an external 3D tracker.

\subsection{Geometric Analysis of Calibration Targets}
\label{sec:geometry}
We first examine how the geometry of commonly used calibration targets, which includes spheres, cones, and wires, affects the constraints imposed on the unknown US-to-robot transformation.
We formulate calibration as a parameter-estimation problem and analyze the resulting constraints through the information matrix of their residuals.

Let $O_b$, $O_{ee}$, and $O_{us}$ denote the robot-base, end-effector, and US-image frames. For the $k$-th scan, forward kinematics provides the \emph{known} EE pose $\mathbf{T}^{b}_{ee,k}=(\mathbf{R}^{b}_{ee,k},\mathbf{t}^{b}_{ee,k})\in \mathrm{SE}(3)$, while the US-to-EE calibration $\mathbf{T}^{ee}_{us}=(\mathbf{R}^{ee}_{us},\mathbf{t}^{ee}_{us})$ is the \emph{unknown} we seek. Detecting the target in image $k$ yields a point $\mathbf{p}^{us}_{k}\in\mathbb{R}^{3}$, reconstructed into the base frame:
\begin{equation}
\mathbf{x}_k(\boldsymbol{\theta})=\mathbf{R}^{b}_{ee,k}\!\left(\mathbf{R}^{ee}_{us}\,\mathbf{p}^{us}_{k}+\mathbf{t}^{ee}_{us}\right)+\mathbf{t}^{b}_{ee,k},
\label{eq:fwd}
\end{equation}
where $\boldsymbol{\theta}\in\mathbb{R}^{6}$ parameterizes the rotational and translational components of the unknown transformation $\mathbf{T}^{ee}_{us}$.

Under a \emph{point-convergence} constraint, target observations reconstructed from different robot poses should converge to a common but unknown target location.
In contrast, a wire-based constraint permits reconstructed points to vary along the wire direction.
To compare these constraints, we characterize the sensitivity of the reconstructed point $\mathbf{x}_k$ to the six calibration parameters $\boldsymbol{\theta}$ through the $3\times6$ Jacobian matrix $\mathbf{G}_k=\partial\mathbf{x}_k/\partial\boldsymbol{\theta}$.
As the target location is itself unknown, a displacement shared across all scans can be absorbed into this unknown location and thus provides no information about the calibration parameters.
We therefore subtract the average Jacobian matrix
$\bar{\mathbf{G}}=\frac{1}{N}\sum_{j=1}^{N}\mathbf{G}_j$ over all $N$ scans and use $\tilde{\mathbf{G}}_k=\mathbf{G}_k-\bar{\mathbf{G}}$ as:
\begin{equation}
\begin{aligned}
\tilde{\mathbf{H}}_{\mathrm{point}} =\sum_{k}\tilde{\mathbf{G}}_{k}^{\top}\tilde{\mathbf{G}}_{k},~
\tilde{\mathbf{H}}_{\mathrm{line}} =\sum_{k}\tilde{\mathbf{G}}_{k}^{\top}\mathbf{P}^{\perp}\tilde{\mathbf{G}}_{k}.
\end{aligned}
\label{eq:target_information}
\end{equation}
Here, $\tilde{\mathbf{H}}_\bullet$ is the information matrix of $\boldsymbol{\theta}$, and $\mathbf{P}^\perp = \mathbf{I}_3 - \hat{\mathbf{u}}\hat{\mathbf{u}}^\top$ removes the component along the wire direction $\hat{\mathbf{u}}$.

The sphere retains this component, so it contains all of the wire's information plus the direction discarded by the wire. 
The sphere thus constrains at least as many calibration directions, provided that the probe poses vary across scans. 
A cone tip gives the same point constraint in principle, but it is reliably visible only near the apex and at a limited range of viewing angles. 
Thus, a sphere can be localized from more directions, making it easier to obtain the pose diversity required for a well-conditioned calibration.

\subsection{Multi-Pose Sphere Acquisition}
\label{sec:conditioning}
The geometric analysis in Sec.~\ref{sec:geometry} motivates acquiring the spherical fiducial from diverse probe orientations and image positions while maintaining the target within the imaging field.
We achieve this using a fixed multi-pose acquisition protocol about a single spherical fiducial (Fig.~\ref{fig:traj}).

From a nominal pose with the probe facing the sphere, we apply rotations of $\pm10^\circ$ about each probe axis ($x,y,z$) and in-plane translations of $\pm3$~mm. 
To achieve all three rotational degrees of freedom while bounding the number of scans, the rotation set is the nominal pose plus the eight $(\pm10^\circ,\pm10^\circ,\pm10^\circ)$ corners ($9$ orientations), each crossed with the nominal and four $(\pm3,\pm3)$~mm translation corners ($5$ positions), for a total of $9\times5=45$ poses. 
At every pose the sphere stays within the imaging field, and the resulting orientation and lever-arm diversity renders the $6$-DOF calibration well-conditioned per Sec.~\ref{sec:geometry}. 
The US-to-EE transform $\mathbf{T}^{ee}_{us}$ is needed to convert this trajectory into robot commands, but it is unknown before calibration. 
We therefore generate the robot poses using a coarse initial estimate, typically obtained from the 3D-printed holder's CAD model or a caliper measurement and accurate to within $\pm3$~mm and $\pm10^\circ$ in every direction. 
The $45$ poses are then executed automatically, providing the US sweeps used for spherical-fiducial localization.

\begin{figure*}[t]
\centering
\includegraphics[width=0.90\textwidth]{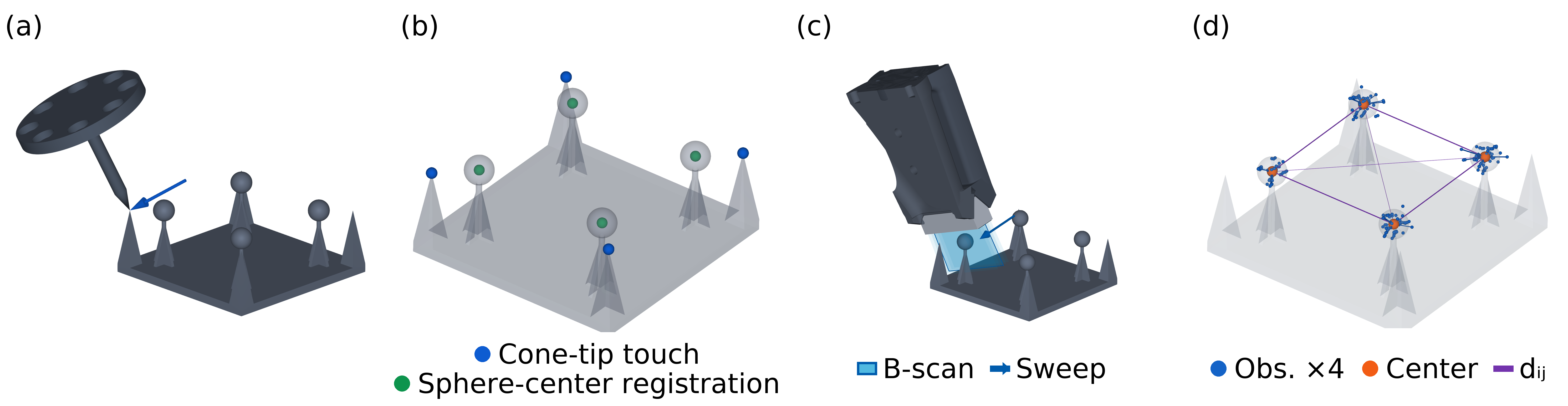}
\caption{Calibration procedure. (a) Cone-tip positions are localized by touch measurements. (b) The phantom model is registered to the robot base frame, providing reference sphere center positions. (c) Each sphere is scanned from multiple probe orientations using US probe. (d) Estimated sphere center locations by US scan. Convergence and known inter-sphere geometry are used.}
\label{fig:process}
\end{figure*}

\subsection{Reference Target Localization and Kinematics Refinement}
\label{sec:touch_localization}

To independently evaluate calibration accuracy, we obtain reference sphere-center locations using the touch-reference principle of Oliveira et al.~\cite{oliveira2025calibration}.
For this measurement, the probe holder is replaced by the touch stylus shown in Fig.~\ref{fig:process}(a).
Three independent datasets (Sets A--C) were acquired in separate sessions, each containing touch measurements of four cone vertices and multi-pose US sweeps of four spheres.
For each dataset under evaluation, the UR3's Denavit--Hartenberg (DH) parameters are refined using touch measurements from the other two datasets and then held fixed.
The stylus-tip offset and phantom pose are subsequently estimated from the current dataset to provide reference sphere centers used exclusively for accuracy evaluation and not for US calibration.
Reconstruction accuracy is further evaluated on cone and triangular-prism phantoms by aligning the reconstructed point clouds with their CAD models and reporting the post-registration geometric residual.

Touch processing has two steps. First, estimate a shared DH model from the other two datasets. Second, freeze that model and fit the tool-tip convergence in the current dataset to obtain its accuracy reference.  

\begin{figure*}[t]
\centering
\vspace{2mm}
\includegraphics[width=0.90\textwidth]{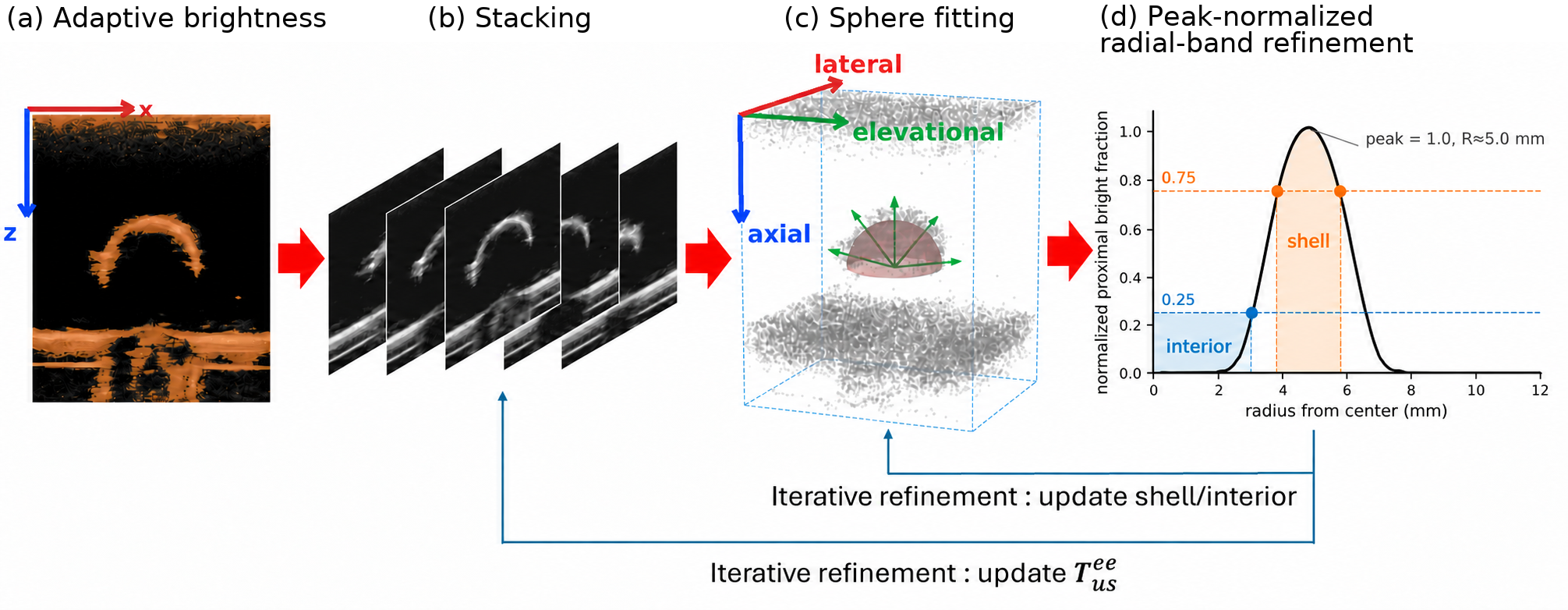}
\caption{Threshold-free sphere localization algorithm: (a)~adaptive brightness, (b)~stacking the elevational sweep into a 3D volume, (c)~count-based sphere fitting within the search ROI, with the dashed box showing the ROI and the red surface showing the fitted proximal hemisphere, and (d)~peak-normalized radial-band refinement. The interior band extends from the center to the first rising crossing of 0.25, while the shell band is bounded by the rising and falling crossings of 0.75. The arrows mark iterative refinement of the shell/interior bands and the calibration $\mathbf{T}^{ee}_{us}$ transform.}
\label{fig:localization}
\end{figure*}

\subsubsection{DH table refinement (Step 1)}
In each training dataset $r$, the stylus touches each of the four cone vertices six times. 
Let $\mathbf{u}_r \in \mathbb{R}^3$ be the unknown tip offset in the end-effector frame and $Y_r=(R_r,\mathbf{t}_r)$ the unknown phantom-to-base pose.
The corresponding model vertices are $\mathbf v_{rj}=R_r\mathbf c_j^{\mathrm{CAD}}+\mathbf t_r$, where $\mathbf c_j^{\mathrm{CAD}}$ is the known CAD vertex. Forward kinematics with scaled DH corrections $\mathbf z$ maps $\mathbf u_r$ to the measured base-frame tip position $\mathbf p^{\mathrm{tip}}_{rjk}(\mathbf z,\mathbf u_r)$. In the first step, we jointly estimate the shared DH corrections and each training dataset's tip and phantom pose by minimizing
\begin{equation}
\begin{aligned}
\mathcal L_{\mathrm{touch}}(\mathbf z,\{\mathbf u_r,Y_r\})
={}&\sum_{r\in\mathcal D_{\mathrm{train}}}\sum_{j=1}^{4}\sum_k
\|\mathbf p^{\mathrm{tip}}_{rjk}-\mathbf v_{rj}\|_2^2\\
&+0.01\|\mathbf z\|_2^2.
\end{aligned}
\label{eq:touch_loss}
\end{equation}
This makes repeated touches converge at each vertex while the common rigid pose preserves the CAD inter-vertex distances. No separate CAD-registration is added. Position residuals are in millimetres. Angular and length corrections in $\mathbf z$ are scaled by $1^\circ$ and $1$~mm, respectively. Bounded nonlinear least squares jointly updates all unknowns, retaining the lowest training loss from 12 deterministic initializations. The Trust-Region Reflective (TRF) algorithm is employed here and for all least-squares fitting in the remainder of the paper. The DH regularizer is applied once across the two training datasets to discourage excessive departures from nominal kinematics. Its coefficient of $0.01$ was fixed across folds, rather than optimized using the held-out US errors.

\subsubsection{CAD-guided reference target (Step-2)}
Let $\widehat{\mathbf z}$ denote the DH corrections learned from the other two datasets. For the current dataset, we keep $\widehat{\mathbf z}$ fixed and estimate only its tip offset $\mathbf u$ and phantom pose $Y=(R,\mathbf t)$ by minimizing
\begin{equation}
\mathcal L_{\mathrm{ref}}(\mathbf u,Y)
=\sum_{j=1}^{4}\sum_k
\|\mathbf p^{\mathrm{tip}}_{jk}(\widehat{\mathbf z},\mathbf u)-\mathbf v_j(Y)\|_2^2,
\label{eq:touch_reference}
\end{equation}
where $\mathbf v_j(Y)=R\mathbf c_j^{\mathrm{CAD}}+\mathbf t$. Thus, the current dataset's repeated touches converge to the four rigidly related CAD vertices without updating DH. The fitted pose maps the CAD sphere centers into the robot-base frame to provide the accuracy reference (procedure in Fig.~\ref{fig:process} (a) and (b)).

\subsection{Threshold-Free Sphere Localization}
\label{sec:localization}

The multi-pose acquisition provides volumetric US sweeps of each spherical fiducial across different probe orientations and image positions.
For each sweep, the sphere center is automatically localized from the US intensity distribution without a manually defined intensity threshold.
Localization is performed in the lateral, axial, and elevational sweep coordinates, with the resulting center mapped to the robot-base frame only during calibration.
The sphere radius $R$ is the only target-specific geometric prior, with all detector settings fixed across datasets (Fig.~\ref{fig:localization}).

\subsubsection{Adaptive brightness} 
For each B-scan $f$, we compute its mean intensity $\mu_f$ and classify samples above $\mu_f$ as bright (Fig.~\ref{fig:localization}(a)).  Because $\mu_f$ is recomputed for every frame, no user-selected intensity cutoff is required.  Bright samples from speckle, reverberation, and other phantom structures are rejected by the geometric scoring below.

\subsubsection{Sweep volume} The B-scans are stacked at the robot-measured spacing to form an initial volume (Fig.~\ref{fig:localization}(b)).  After calibration updates $\mathbf{T}^{ee}_{us}$, the same B-scans are stacked once more using the updated transform to correct their relative alignment, and sphere localization is repeated.  Only the probe-facing hemisphere is sampled because the opposite surface is obscured by acoustic shadow.

\subsubsection{Count-based exhaustive search} 
Candidate centers $\mathbf{c}$ are searched on a $1$-mm grid over the lateral, axial, and elevational extent of the sweep, beginning one radius $R$ below the image surface to keep the sphere within the imaging region.
At each candidate, near-uniform Fibonacci directions sample a proximal shell at radius $R$ and a set of volume-distributed points inside the proximal half-ball.  
At each off-grid sample location, the intensity of the nearest voxel is used and compared with the mean intensity of that voxel's B-scan. The candidate is scored by
\begin{equation}
s(\mathbf{c}) = f_{\mathrm{shell}}(\mathbf{c}) - f_{\mathrm{interior}}(\mathbf{c}),
\end{equation}
where $f_{\mathrm{shell}}$ and $f_{\mathrm{interior}}$ are the fractions of in-bounds samples brighter than their corresponding frame means.  
A correct center maximizes the bright-shell and dark-interior contrast, yielding the seed $\arg\max_{\mathbf{c}}s(\mathbf{c})$.
Samples falling outside the acquired volume are excluded from both the numerator and denominator of these fractions, allowing partially imaged caps to be scored from their visible portion.

\subsubsection{Radial-band refinement} 

For each sphere and trial radius, a probe-facing hemispherical surface is sampled around the current center across all $45$ sweep volumes (Fig.~\ref{fig:localization}(d)).
The fraction of valid samples brighter than their corresponding B-scan means is averaged across poses to form a peak-normalized radial brightness profile.
The first rising crossing of $0.25$ defines the interior limit, while the $0.75$ crossings define the shell band.
These bands replace the initial samples for two refinement passes.

\begin{figure}[t]
\centering
\includegraphics[width=0.90\columnwidth]{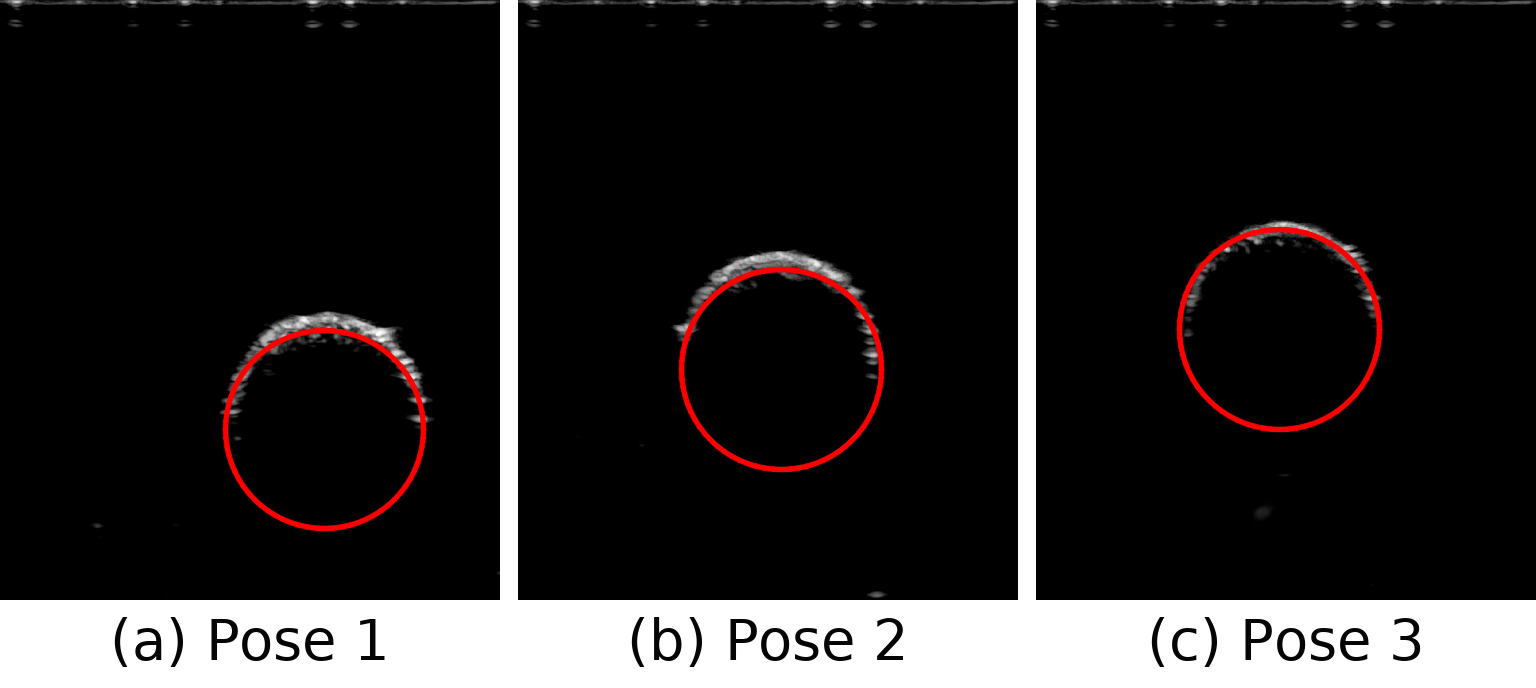}
\caption{Representative B-mode cross-sections acquired at three robot poses: (a) an XYZ Euler-angle offset of $(-10^\circ,+10^\circ,+10^\circ)$ with an in-plane offset of $(+3,+3)$~mm, (b) the same orientation at the centered in-plane position, and (c) an XYZ Euler-angle offset of $(+10^\circ,+10^\circ,+10^\circ)$ at the centered position. All rotations and translations are defined relative to the nominal probe pose. The red contours denote the sphere boundaries predicted by the automatic localization method.}
\label{fig:detected_cross_sections}
\end{figure}

The final grid point is the localized sphere center for that sweep.  
After each calibration update, rotation-dependent shear correction and localization are repeated as described in Sec.~\ref{sec:Framework}, with representative detections shown in Fig.~\ref{fig:detected_cross_sections}.

\subsection{Single- and Three-Marker Calibration}
\label{sec:Framework}
Using the sphere centers localized from the multi-pose US sweeps, we estimate the US-to-EE transformation by enforcing geometric consistency among repeated observations of the spherical fiducials.
As a multi-point geometric reference, three non-collinear markers provide the minimum configuration for defining a rigid reference frame, with one marker establishing the origin and the remaining two determining independent directions~\cite{zhang2018phantom}.
Such multi-point geometric constraints have also been employed in US calibration, where three or more fiducials with known relative 3D locations provide spatial constraints for estimating the calibration transformation~\cite{cheng2017active,yang2025robot}.
We therefore use three-marker calibration as a multi-fiducial reference formulation, in which the known relative geometry among the three sphere centers constrains the US-to-EE transformation.
In contrast, the proposed single-marker formulation removes the requirement for an explicitly defined multi-fiducial frame and instead exploits repeated observations of a single spherical fiducial from diverse robot poses.

\subsubsection{Generalized multi-markers calibration}
Let $X=\mathbf T^{ee}_{us}$ be the unknown US-to-EE transform and $Y=(R,\mathbf t)=\mathbf T^b_p$ the unknown phantom-to-base transform. 
The three sphere centers satisfy $\mathbf C_s=R\mathbf c_s^{\mathrm{CAD}}+\mathbf t$, where $\mathbf c_s^{\mathrm{CAD}}$ is the known center in the phantom frame.
Estimating a common rigid pose preserves the known inter-sphere distances without requiring the sphere centers to be pre-localized in the robot-base frame. Touch-derived reference coordinates are excluded from this fit. 
For each sphere $s$, we use Eq.~\eqref{eq:fwd} to reconstruct its observed centers $\mathbf p_{sk}(X)$ in the robot-base frame. 
Let $\overline{\mathbf p}_s(X)$ denote the centroid of the $n_s$ retained observations ($n_s=45$ in our experiments). The generalized US calibration loss for both single-marker and three-marker calibration methods is:
\begin{equation}
\begin{aligned}
\mathcal{L}_{\mathrm{US}}(X,Y)
={}&
\sum_{s\in\mathcal{S}_{\mathrm{train}}}\sum_k
\left\|
\mathbf{p}_{sk}(X)-\overline{\mathbf{p}}_s(X)
\right\|_2^2 \\
&+
\sum_{s\in\mathcal{S}_{\mathrm{train}}} n_s
\left\|
\overline{\mathbf{p}}_s(X)-\mathbf{C}_s
\right\|_2^2 .
\end{aligned}
\label{eq:distance_objective}
\end{equation}
Building upon the three-marker formulation, we also estimate the calibration from repeated observations of a single sphere. 

\subsubsection{Single-marker calibration}
For single sphere calibration, center-geometry error becomes trivially 0, since geometry error cannot be defined for single sphere center.
Because localization stacks the elevational sweep under an assumed calibration (Sec.~\ref{sec:localization}), a rotational error in $\mathbf T^{ee}_{us}$ shears the reconstructed sphere and biases its localized center. 
To mitigate this coupling, localization and calibration are alternated: after each calibration update, the B-scans are re-stacked using the updated $\mathbf{T}_{us}^{ee}$ to restore their relative alignment, and the sphere center is re-localized in the corrected volume.

\subsubsection{Evaluation protocol}
\label{eval:metric}
To evaluate the proposed single-marker against the three-marker calibration under consistent experimental conditions, we use complementary sphere-wise calibration and test splits with a cross-validation design to evaluate whether the estimated calibration generalizes to spatially distinct sphere locations not used during calibration.
For each dataset, four complementary splits are evaluated with DH parameters estimated from the other two datasets and held fixed.
In each split, single-marker calibration uses one sphere for calibration and the remaining three for evaluation, while three-marker calibration uses the complementary three spheres and evaluates on the remaining sphere.
Touch-derived references are used only for accuracy evaluation.

\vspace{-1mm}
\section{Experiments and Results}
\label{sec:results}

\subsection{Calibration Precision and Accuracy}

Using the evaluation protocol described in Sec.~\ref{eval:metric}, we first assess the precision and accuracy of the proposed single-marker calibration across the three experimental datasets. Across 12 acquisitions, active data collection required $349.5\pm7.1$~s per sphere (mean $\pm$ SD), corresponding to approximately 5.8 and 17.5~min for single- and three-marker calibration.
For each dataset, the sphere precision is the mean Euclidean distance of its retained reconstructed centers from their centroid. 
Fig.~\ref{fig:detected_cross_sections} shows that the detected sphere boundaries in representative cross-sections closely follow the bright contours of the spheres' upper hemispheres. 
Tool-point precision is computed analogously from the six different robot poses at each cone vertex. 

Sphere accuracy is the Euclidean distance between each reconstructed centroid and its touch-derived reference center, obtained from the joint tip/phantom-pose fit with DH parameters frozen. For each dataset, the DH parameters learned from the other two datasets were held fixed throughout evaluation. Across the three leave-one-dataset-out DH models, the RMS of changes from factory reported DH parameters in $(\theta,d,a,\alpha)$ were $(2.51^\circ, 0.24~\mathrm{mm}, 1.51~\mathrm{mm}, 0.27^\circ)$, respectively. 
Since the tool points define this reference, only their precision is reported. 
For sphere metrics, each fold averages the three test-sphere errors. 
Each set reports the mean and sample standard deviation (SD) of its four fold averages, and the overall row uses all 12 folds. 
Tool precision is summarized across the four touch points and six poses for each point per dataset. Table~\ref{tab:calibration_cv} summarizes the four-fold evaluation of single-marker calibration for each dataset.

\begin{table}[h]
\caption{Single-marker calibration errors (mm).}
\label{tab:calibration_cv}
\centering
\renewcommand{\arraystretch}{1.15}
{\scriptsize
\setlength{\tabcolsep}{1pt}
\begin{tabular*}{\columnwidth}{@{\extracolsep{\fill}}llcccc@{}}
\hline
Metric & Statistic & A & B & C & Overall \\
\hline
Tool prec.
& MAE$\pm$SD
& $0.60\pm0.04$
& $0.53\pm0.11$
& $0.70\pm0.13$
& $0.61\pm0.12$ \\

& RMSE
& $0.60$
& $0.53$
& $0.70$
& $0.62$ \\

Sphere prec.
& MAE$\pm$SD
& $0.71\pm0.02$
& $0.89\pm0.02$
& $0.96\pm0.02$
& $0.85\pm0.11$ \\

& RMSE
& $0.71$
& $0.89$
& $0.96$
& $0.86$ \\

Sphere acc.
& MAE$\pm$SD
& $1.16\pm0.12$
& $2.26\pm0.14$
& $1.82\pm0.30$
& $1.75\pm0.51$ \\

& RMSE
& $1.16$
& $2.27$
& $1.84$
& $1.81$ \\
\hline
\end{tabular*}

\vspace{2pt}
\parbox{\columnwidth}{\scriptsize
\textit{Note:} For tool precision, one sample is obtained for each of the four cone vertices by averaging the distances of six touch localizations from their vertex-specific centroid. Thus, each set contains four tool-precision samples, and the overall column contains 12 samples. For each sphere metric, one sample is obtained per single-marker split by averaging the corresponding errors of the three held-out spheres. Each set contains four samples, and overall contains all 12 samples. SD, mean absolute error (MAE) and root mean square error (RMSE) use the same samples.
}
}
\end{table}

Table~\ref{tab:three_marker_cv} summarizes the four-fold leave-one-sphere-out evaluation of three-marker calibration for each dataset. 

\begin{table}[h]
\caption{Three-marker calibration errors (mm).}
\label{tab:three_marker_cv}
\centering
\renewcommand{\arraystretch}{1.15}
{\scriptsize
\setlength{\tabcolsep}{1pt}
\begin{tabular*}{\columnwidth}{@{\extracolsep{\fill}}llcccc@{}}
\hline
Metric & Statistic & A & B & C & Overall \\
\hline
Sphere prec.
& MAE$\pm$SD
& $0.70\pm0.07$
& $0.88\pm0.02$
& $0.95\pm0.07$
& $0.84\pm0.12$ \\

& RMSE
& $0.70$
& $0.88$
& $0.95$
& $0.85$ \\

Sphere acc.
& MAE$\pm$SD
& $1.13\pm0.20$
& $2.24\pm0.15$
& $1.80\pm0.26$
& $1.72\pm0.51$ \\

& RMSE
& $1.14$
& $2.25$
& $1.81$
& $1.79$ \\
\hline
\end{tabular*}

\vspace{2pt}
\parbox{\columnwidth}{\scriptsize
\textit{Note:} Precision and accuracy are defined as in Table~\ref{tab:calibration_cv}. Unlike the single-marker evaluation, each
sample is the result for one held-out sphere in a leave-one-sphere-out fold (four samples per set). SD and RMSE use these same samples.
}
}
\end{table}

\subsection{Volumetric Reconstruction of Different Phantoms}

\begin{figure}[htp]
\centering
\includegraphics[width = 0.90\columnwidth]{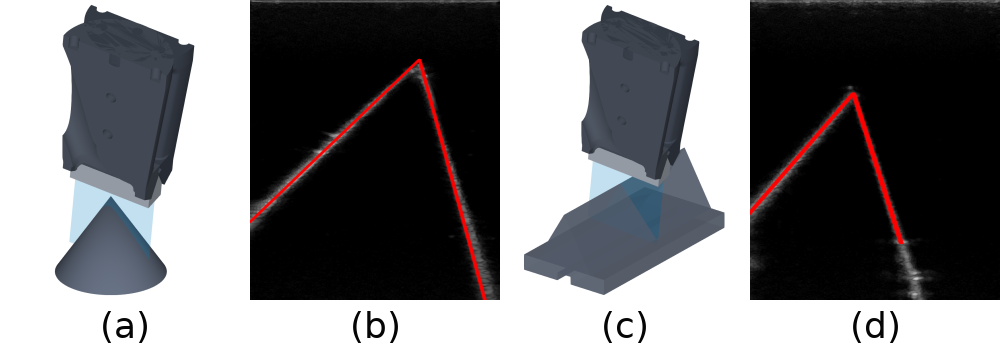}
\caption{(a) Cone phantom scanned from a tilted direction. (b) US image of the scanned cone phantom and fitted red lines to the bright boundaries. (c) Triangular prism phantom scanned from tilted direction. 
(d) Corresponding US image and fitted red lines to the bright boundaries.}
\label{fig:cross_phantom_detection}
\end{figure}
\begin{table*}[t]
\vspace{2mm}
\caption{Single- and Three-Marker Point-to-Surface Errors After ICP
Registration (mm)}
\label{tab:cross_phantom_icp}
\centering
\renewcommand{\arraystretch}{1.15}
\begin{tabular*}{\textwidth}{@{\extracolsep{\fill}}llcccccc@{}}
\hline
& & \multicolumn{3}{c}{Single-marker (Sphere~1)}
& \multicolumn{3}{c}{Three-marker (Spheres~1--3)} \\
\cline{3-5}\cline{6-8}
Phantom & Set
& MAE $\pm$ SD & RMSE & Max
& MAE $\pm$ SD & RMSE & Max \\
\hline
Cone & A
& $0.36\pm0.30$ & $0.46$ & $1.99$
& $0.34\pm0.29$ & $0.45$ & $1.99$ \\
Cone & B
& $0.45\pm0.39$ & $0.60$ & $3.04$
& $0.44\pm0.39$ & $0.59$ & $2.75$ \\
Cone & C
& $0.43\pm0.35$ & $0.56$ & $2.34$
& $0.41\pm0.34$ & $0.53$ & $2.42$ \\
Cone & Overall
& $0.41\pm0.35$ & $0.54$ & $3.04$
& $0.40\pm0.35$ & $0.53$ & $2.75$ \\
\hline
Triangular prism & A
& $0.45\pm0.34$ & $0.57$ & $1.59$
& $0.44\pm0.33$ & $0.55$ & $1.50$ \\
Triangular prism & B
& $0.46\pm0.44$ & $0.64$ & $1.96$
& $0.44\pm0.42$ & $0.61$ & $1.89$ \\
Triangular prism & C
& $0.57\pm0.44$ & $0.72$ & $1.97$
& $0.56\pm0.42$ & $0.70$ & $1.86$ \\
Triangular prism & Overall
& $0.50\pm0.41$ & $0.64$ & $1.97$
& $0.48\pm0.40$ & $0.62$ & $1.89$ \\
\hline
\end{tabular*}

\vspace{2pt}
\parbox{\textwidth}{\footnotesize
\textit{Note:} Sets A--C use the corresponding fixed leave-one-dataset-out DH models. Single-marker calibration uses Sphere~1, whereas three-marker calibration uses Spheres~1--3. MAE, SD, RMSE, and Max are computed over point-to-CAD-surface distances. Overall rows concatenate Sets A--C.
}
\end{table*}

To evaluate generalization beyond the spherical calibration fiducials, we assess volumetric reconstruction on geometrically distinct cone and triangular-prism phantoms (Fig.~\ref{fig:cross_phantom_detection}).
Because the probe was remounted after the calibration experiments, $\mathbf T^{ee}_{us}$ was re-estimated from a separate scan of the four-sphere phantom using the same single- and three-marker procedures. For each of the three frozen DH models, a single-marker fit using sphere 1 and three-marker fit using spheres 1--3 were used to estimate the US-to-EE transform. Each resulting DH/US-transform pair was then applied to both geometric phantoms. The cone measures $60$~mm in base diameter and $44.64$~mm in height, whereas the triangular prism measures $60$~mm in width, $120$~mm along its upper ridge, and $44.64$~mm in height. The validation trajectories followed the same pose-diversity principle as the sphere protocol. For the cone, 45 linear sweeps were acquired around the apex, as it did around sphere center, with each starting pose selected so that the apex remained within the acquired B-mode sequence. For the triangular prism, the same 45 different probe positions and orientations were defined around the starting apex point. All 45 linear trajectories were mutually parallel and followed the upper ridge. Their travel direction was held fixed, while the starting position and probe orientation were varied to reconstruct the same ridge from multiple views. At each pose, the probe performs a $\pm1$~cm linear sweep and acquires $21$ frames at $1$~mm spacing, as it does for sphere.

Each B-mode frame from both phantoms was segmented using Otsu thresholding. The resulting binary mask defined candidate regions for the two boundaries. For each lateral row, the actual maximum-intensity pixel near each fitted boundary was retained as a boundary sample. The central 10 of the 21 frames in each trajectory were used. For each set and calibration method, the corresponding US-to-EE transform and joint-touch DH model were held fixed during reconstruction. The reconstructed boundary points were subsequently transformed into the robot-base frame and rigidly registered to the corresponding CAD surface using point-to-point iterative closest point (ICP) algorithm~\cite{besl1992method}.

\begin{figure*}[t]
\centering
\vspace{2mm}
\includegraphics[width=0.90\textwidth]{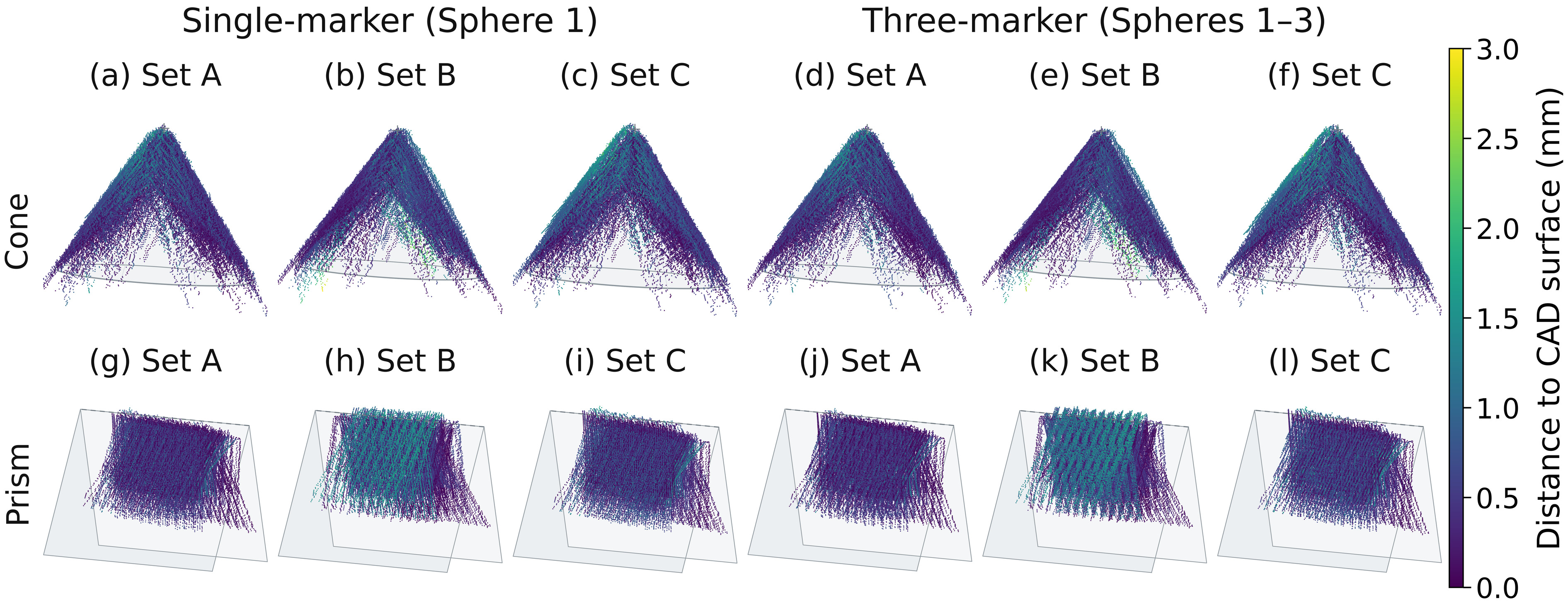}
\caption{Point-to-surface error heatmaps after ICP registration, shown as orthographic top-down views. The top row (a--f) shows the cone phantom, and the bottom row (g--l) shows the triangular-prism phantom. Within each row, the first three panels show the Sphere~1 single-marker calibration results for Sets A--C, while the last three show the Spheres~1--3 three-marker calibration results for Sets A--C.}
\label{fig:phantom_heatmap}
\end{figure*}

Fig.~\ref{fig:phantom_heatmap} shows point-to-surface error heatmaps. Table~\ref{tab:cross_phantom_icp} reports the corresponding quantitative results. Pooling point-to-surface distances across Sets A--C, the single-marker calibration yielded MAE$\pm$SD values of $0.41\pm0.35$~mm for the cone and $0.50\pm0.41$~mm for the triangular prism, with corresponding RMSEs of $0.54$ and $0.64$~mm. The three-marker calibration yielded $0.40\pm0.35$ and $0.48\pm0.40$~mm, with RMSEs of $0.53$ and $0.62$~mm, respectively.

\section{Discussion}
\label{sec:discussion}

In this work, we developed a tracker-free robotic US calibration framework that combines robot forward kinematics with automated, threshold-free localization of spherical fiducials.
The proposed single-marker formulation estimates the US-to-EE transformation, $\mathbf{T}^{ee}_{us}$, from repeated multi-pose observations of a single passive sphere, eliminating the need for an external 3D tracker or a predefined multi-fiducial reference frame.
To assess the feasibility of single-marker calibration, we compare it against a three-marker baseline that additionally exploits the known relative geometry of three non-collinear sphere centers.
Across both formulations, pose-diverse US sweeps are processed using the same threshold-free localization pipeline, which adapts to the intensity distribution of each B-scan without manual contouring or a manually tuned absolute intensity threshold.

Single-marker calibration produced sphere-center accuracy and precision errors of $1.75\pm0.51$~mm and $0.85\pm0.11$~mm. 
Three-marker calibration produced $1.72\pm0.51$~mm and $0.84\pm0.12$~mm, respectively. Both sets of results are numerically close to the 1.72-mm accuracy and 0.80-mm precision reported by Oliveira et al.~\cite{oliveira2025calibration}. 
Using the corresponding fixed joint-touch DH models, the Sphere~1 single-marker calibration produced pooled point-to-surface MAE$\pm$SD values of $0.41\pm0.35$~mm for the cone and $0.50\pm0.41$~mm for the triangular prism, closely matching the Spheres~1--3 three-marker results of $0.40\pm0.35$~mm and $0.48\pm0.40$~mm, respectively. These residuals quantify reconstructed shape consistency. 
These results show that the estimated transformation is sufficiently consistent to reconstruct unseen geometries at submillimeter scale.

Several factors may contribute to the remaining error relative to the independent reference, including uncertainty from manual stylus contact and remounting, US image resolution ($\sim0.7$~mm lateral and $\sim0.35$~mm axial), acoustic-interface and speed-of-sound biases, robot kinematic/repeatability errors, and 3D-printing tolerances. 
The $1.75$-mm sphere accuracy corresponds to approximately $2.5$ lateral resolution cells, while the $0.85$-mm precision approaches one lateral resolution cell. Thus, the measured accuracy reflects the end-to-end system performance rather than the isolated calibration error.

\section{Conclusion}
\label{sec:conclusion}

In summary, we present a tracker-free and automated robotic US calibration framework that eliminates external tracking and its associated hardware, coordinate transformations, and deployment constraints. 
Threshold-free sphere localization further reduces operator-dependent parameter tuning, while spherical geometry supports pose-diverse observations for single-marker calibration. 
The present study is limited to one robot, probe, US system, and phantom design; future work will evaluate cross-system generalizability and reduce the number of required acquisition poses while maintaining calibration accuracy and conditioning. 
Overall, the proposed framework provides a simple and repeatable approach for estimating $\mathbf{T}^{ee}_{us}$ in robotic US systems.

\section{Acknowledgements}

The authors thank Yilin Ma of the Hybrid Dynamic Robotics Lab for assistance with 3D printing the phantom and Alain Zhou of the Department of Mechanical Engineering at the University of Michigan for assistance with accuracy testing.

\bibliographystyle{IEEEtran}
\bibliography{refs}
\end{document}